\documentclass[conference]{IEEEtran}
\IEEEoverridecommandlockouts

\usepackage{cite}
\usepackage{amsmath,amssymb,amsfonts}
\usepackage{algorithmic}
\usepackage{graphicx}
\usepackage{textcomp}
\usepackage{xcolor}
\usepackage{float}
\def\BibTeX{{\rm B\kern-.05em{\sc i\kern-.025em b}\kern-.08em
    T\kern-.1667em\lower.7ex\hbox{E}\kern-.125emX}}
\begin{document}

\title{Fitting Different Interactive Information: Joint Classification of Emotion and Intention\\
}

\author{
\IEEEauthorblockN{\textsuperscript{} Xinger Li}
\IEEEauthorblockA{\textit{Nanjing University of Science and Technology}\\
Nanjing, China \\
lxenjust@njust.edu.cn}
\\
\IEEEauthorblockN{\textsuperscript{} Bo Huang}
\IEEEauthorblockA{\textit{Nanjing University of Science and Technology}\\
Nanjing, China \\
huangbo@njust.edu.cn}

\and

\IEEEauthorblockN{\textsuperscript{} Zhiqiang Zhong}
\IEEEauthorblockA{\textit{Nanjing University of Science and Technology} \\
Nanjing, China \\
1533534827@qq.com}
\\

\IEEEauthorblockN{\textsuperscript{} Yang Yang}
\IEEEauthorblockA{\textit{Nanjing University of Science and Technology}\\
Nanjing, China \\
yyang@njust.edu.cn}
}

\maketitle

\begin{abstract}
This paper is the first-place solution for ICASSP MEIJU@2025 Track I, which focuses on low-resource multimodal emotion and intention recognition. How to effectively utilize a large amount of unlabeled data, while ensuring the mutual promotion of different difficulty levels tasks in the interaction stage, these two points become the key to the competition. In this paper, pseudo-label labeling is carried out on the model trained with labeled data, and samples with high confidence and their labels are selected to alleviate the problem of low resources. At the same time, the characteristic of easy represented ability of intention recognition found in the experiment is used to make mutually promote with emotion recognition under different attention heads, and higher performance of intention recognition is achieved through fusion. Finally, under the refined processing data, we achieve the score of 0.5532 in the Test set, and win the championship of the track.
\end{abstract}


\section{Introduction}
With the continuous development of the multi-modal field, we pay increasingly attention to the complementary information between different modals to achieve more superior task performance\cite{b7}\cite{b4}. In MEIJU25 Challenge@ICASSP2025 competition, Video-based image, text and audio multi-modal emotion and intention recognition tasks also try to use information interaction among different tasks to achieve better classification performance\cite{b1}\cite{b10}. The purpose of Chinese language track I MEIJU25 is to explore the interaction between emotion recognition intention recognition tasks with a few annotated data, so as to improve their respective performance.

Our main contributions of this paper are as follows:
\begin{itemize}
\item Experiment find intention recognition is more easily represented by emotional features, and the performance of intention recognition can be improved through the interaction of task information under different self-attention multi-heads.

\item The use of high confidence pseudo-labels and refined text punctuation improves the performance of each task.

\item The global average embedding of each modal is a better representation of the overall emotion and intent features of the video.
\end{itemize}

Finally, we integrate the reasoning results of intent recognition under different multi-head to achieve complementary intent performance. We believe that the interaction between emotion and intent recognition with different multi-head have greatly improved classification accuracy.

\section{Method}
Figure 1 shows the overall architecture of our solution. Inspired by Liu, Fan et al. \cite{b2}, the final model mainly consists of a multi-modal feature fusion module based on multi-head self-attention and gated mechanism interaction module for the mutual promotion of different difficulty tasks. The details will be explained in subsections D and E.

\begin{figure*}[htb]
  \centering
  \includegraphics[width=1\linewidth]{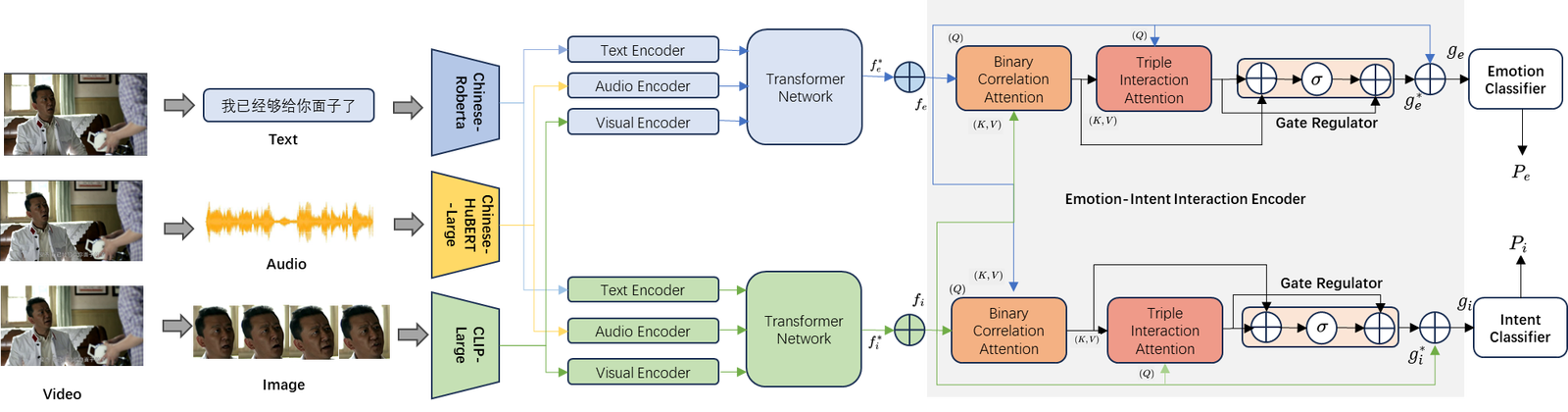}
  \caption{The overall architecture of our solution.}
\end{figure*}


\subsection{Data preprocessing}\label{AA}
Visual: Considering the large number of duplicate pictures in the near moment of the video, we sample 1 picture from the video every 30 frames (about 1s) in order to avoid excessive redundancy and noise.

Audio: By using ffmpeg tool, the original video is extracted into audio with sample rate of 16K, and save into WAV format files.

Text: Although the organizer providing corresponding audio text, manual inspection revealed some errors. Attempts with models like Qwen2.5-7B were unfruitful. We manually corrected a few words and symbols, such as "?", to better classify emotions like Surprise and intents like Questioning.
\subsection{Pseudo-label generation}\label{BB}
In order to make full use of unlabeled data and generate pseudo-labels to expand the dataset, we use the existing labeled data to train the initial model,
and then use the model to predict the unlabeled data and generate pseudo-labels.
Specifically, we set a confidence threshold of 0.99 to select samples and false labels with high confidence, and adopt reasonable sampling strategies to ensure the balance of emotion categories and intention categories\cite{b6}. Finally, these pseudo-label samples
are added to the training set as additional labeled data.
\subsection{Modal processing}\label{CC}
Based on Lian's work and local experiment\cite{b3}, we select CLIP-large, HUBERT-large, and Chinese-Roberta for extracting embeddings from Visual, Audio, and Text modalities, respectively.

For the Visual modality, TalkNet \cite{b5} is used to extract the speaker's face from videos, followed by CLIP-large for embedding extraction. For Audio, we sum the last four hidden layers of the HUBERT-large model to obtain embeddings. For Text, embeddings are extracted for each word and punctuation.

To align modal information and accommodate varying video lengths, we compute the average embedding for each modality as the final representation \cite{b7}\cite{b8}\cite{b9}.

\subsection{Multimodal fusion module}\label{DD}
We employ the same encoding module for both emotion and
intention recognition tasks. For audio and visual embeddings,
we use LSTM encoders to extract higher-dimensional features
and apply max pooling to generate low-dimensional, compact and context-specific representations. For text embeddings, we
use a single-layer TextCNN to achieve similar compactness.
Through experiments, we find that these low-dimensional em-
beddings facilitate task interaction and improve performance.
Finally, we fuse embeddings from all three modalities using
a single-layer multi-head Transformer, resulting in stronger
multi-modal representations.

\subsection{Multi-head self-attention, gated mechanism interaction}\label{EE}
We utilize a multi-head self-attention mechanism to capture modal information of different dimensions, with fewer heads and more heads for intent tasks to get different embedding representations. During the interaction process, fusional Emo and Int embeddings serve as the Query respectively, while other embeddings act as the Value and Key to compute the initial interaction result A. This result is further refined through a second interaction step, using A as the Value and Key. A gating mechanism then learns the weight of the second interaction feature, which is combined with the Query to produce the final feature. To utilize the more easier representation ability in intention tasks, we use self-attention with one and two heads. Through fusion, we achieve improved performance while keeping the interaction module unchanged.  

\section{Experiment}
The experimental results are shown in Table 1. Through experiments, we observe two key findings: (1) Emotion and intention recognition are interrelated, but achieving optimal accuracy for both tasks is challenging due to differences in task complexity. (2) Intention recognition is simpler represented. Considering these, we employ multi-head self-attention and modal fusion while adding Gaussian noise to embeddings for enhanced robustness. We adopt a two-stage training strategy: first, we train the model on clean data, and then we fine-tune it using pseudo-labeled data. This approach improves alignment, enriches modal representations, and enhances overall task performance.
\begin{table}[H]
\caption{The experimental results.Note that the scores corresponding to the fifth row are all intent scores.}
\begin{center}
\resizebox{\columnwidth}{!}{%
\begin{tabular}{|c|c|c|c|}
\hline
\textbf{method} & \textbf{\textit{emotion score/intent score}}& \textbf{\textit{intent score}}& \textbf{\textit{overall score}} \\
\hline
baseline & 0.5265 & 0.5051 & 0.5156 \\
+ pesudo label & 0.5305 & 0.5114 & 0.5208 \\
+ pesudo label + 1textcnn & 0.5725 & 0.4831 & 0.5240  \\
+ one attention head & 0.5134 & 0.5132 & 0.5133 \\
+ one attention head + two attention head & 0.5114 & 0.5132 & 0.5353\\
ensemble & 0.5725 & 0.5353 & 0.5532 \\
\hline
\end{tabular}
}
\label{tab1}
\end{center}
\end{table}

\section{Conclusion}

There is a corresponding relationship between emotion recognition and intention recognition, but the learning rate of different tasks is different, which makes it difficult to achieve the best accuracy of each task at the same time. Considering that intention recognition is a task easier to represent, by allowing self-attention of emotion recognition and intention recognition under different heads to interact, intention recognition pays attention to different information, so as to achieve better results than that under a single head. On the refined data set, the final model fuses different head information in the result of intention recognition through two-stage training, and finally achieves the score of 0.5532 in the online test set, achieving the optimal performance of the overall task.

\end{document}